# Geometric Implications of the Naive Bayes Assumption

**Mark A. Peot***
Rockwell International Science Center
444 High Street, Suite 400
Palo Alto, California 94301

peot@kic.com

## Abstract

A Naive (or Idiot) Bayes network is a network with a single hypothesis node and several observations that are conditionally independent given the hypothesis. We recently surveyed a number of members of the UAI community and discovered a general lack of understanding of the implications of the Naive Bayes assumption on the kinds of problems that can be solved by these networks. It has long been recognized [Minsky 61] that if observations are binary, the decision surfaces in these networks are hyperplanes. We extend this result (hyperplane separability) to Naive Bayes networks with m-ary observations. In addition, we illustrate the effect of observation-observation dependencies on decision surfaces. Finally, we discuss the implications of these results on knowledge acquisition and research in learning.

## 1 INTRODUCTION

While it is widely accepted that Naive Bayes models are a weaker representation for classification problems than a more general belief network, the precise nature of the limits on the expressive power of these models is not so widely known in the UAI community. This paper aims to redress this deficiency by presenting (with extensions) several classic results concerning the expressive power of Naive Bayes classifiers. We generalize this result to more general classifiers and discuss the implications of these results on current research on learning and knowledge acquisition.

---

* http://rpal.rockwell.com/~peot. Also at Knowledge Industries; 350 Cambridge Ave, Suite 385; Palo Alto, CA 94306

## 2 NAIVE BAYES

A Naive (or Idiot) Bayes model is a belief network consisting of a single discrete hypothesis node, $\Omega$, and one or more observation nodes, $\overline{O} = \{O_1, ..., O_n\}$, that are conditionally independent given the hypothesis. $\Omega$ can assume any one of $m$ mutually exclusive and collectively values, $\omega_1, ..., \omega_m$. We will assume that each of the $O_i$ are discrete and can assume any one of $k_i$ values, $o_{i1}, ..., o_{ik_i}$. Such a network is illustrated below.

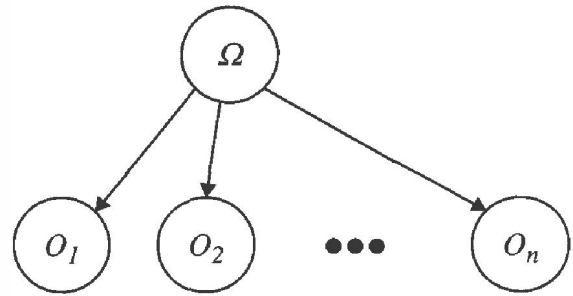

**Figure 1:** A Naive (Idiot) Bayes network.

The probability of the hypotheses, $P\{\Omega = \omega_i | \overline{O} = \bar{o}\}$ is given by

$$P\{\Omega = \omega_i|\overline{O}\} = \frac{P\{\Omega = \omega_i\}}{P\{\overline{O}\}} \prod_{j=1}^{n} P\{O_j|\Omega = \omega_i\} \quad (1)$$

## 3 NAIVE BAYES CLASSIFICATION WITH BINARY OBSERVATIONS

The classification problem is the problem of determining the most likely hypothesis given a set of observations. We wish to determine the hypothesis of maximum posterior probability, that is, we wish to determine $\omega_k$ such that

$P\{\omega_k|\overline{O}\} \geq P\{\omega_i|\overline{O}\}, \forall i$. The technique that we will use is to determine a set of *decision functions* that each determine whether $P\{\omega_i|\overline{O}\} \geq P\{\omega_j|\overline{O}\}$ and examine the *decision surfaces* induced by those decision functions. Minsky [1961] was the first to show that these decision surfaces are hyperplanes when the observations are binary [Duda&Hart, 1973].

A *decision function* is a continuous function $r: (R^n \to R)$ such that

- $r(\overline{O}) > 0$ whenever we prefer classification $\Omega_1$ to $\Omega_2$,
- $r(\overline{O}) < 0$ whenever we prefer $\Omega_2$ to $\Omega_1$, and
- $r(\overline{O}) = 0$ whenever we are indifferent between $\Omega_1$ and $\Omega_2$.

The function $r(X) = 0$ defines a *decision surface*. On one side of this surface, we will prefer $\Omega_1$, and on the other we will prefer $\Omega_2$.

Assume that each of the $O_i$ is binary with value 0 (false) or 1 (true). Each of the possible instantiations of $\overline{O}$ correspond to a vertex of the n-cube $\{0, 1\}^n$.

**Theorem 1:** [Minsky 1961] The decision surfaces in a Naive Bayes network are hyperplanes.

**Proof:** First of all, note that[1]

$$P\{O_j|\Omega = \omega_i\}$$
$$= P\{O_j = 0|\Omega = \omega_i\} \left( \frac{P\{O_j = 1|\Omega = \omega_i\}}{P\{O_j = 0|\Omega = \omega_i\}} \right)^{O_j} \quad (2)$$

Substitute (2) into (1) and take the log to obtain:

$$\log P\{\Omega = \omega_i|\overline{O}\} =$$
$$\log \left( \frac{P\{\Omega = \omega_i\}}{P\{\overline{O}\}} \prod_{k=1}^{n} P\{O_k = 0|\Omega = \omega_i\} \right) \quad (3)$$
$$+ \sum_{j=1}^{n} O_j \log \left( \frac{P\{O_j = 1|\Omega = \omega_i\}}{P\{O_j = 0|\Omega = \omega_i\}} \right)$$

Let

$$w_{i0} = \log \left( P\{\Omega = \omega_i\} \prod_{k=1}^{n} P\{O_k = 0|\Omega = \omega_i\} \right)$$

and

$$w_{ij} = \log \left( \frac{P\{O_j = 1|\Omega = \omega_i\}}{P\{O_j = 0|\Omega = \omega_i\}} \right)$$

Thus

$$\log P\{\Omega = \omega_i|\overline{O}\} = w_{0i} + \sum_{j=1}^{n} w_{ij} O_j - \log(P\{\overline{O}\})$$

Define the decision function

$$r_{ab}(\overline{O}) = \log P\{\Omega = \omega_a|\overline{O}\} - \log P\{\Omega = \omega_b|\overline{O}\}$$
$$= (w_{a0} - w_{b0}) + \left( \sum_{j=1}^{n} (w_{aj} - w_{bj}) O_j \right)$$

---

1. Remember that $O_i$ is either 0 or 1.

Now, the probability of $\omega_a$ is higher than the probability of $\omega_b$ whenever $r_{ab}(\overline{O}) > 0$.

The decision functions $r_{ab}(\overline{O})$ are, of course, hyperplanes.

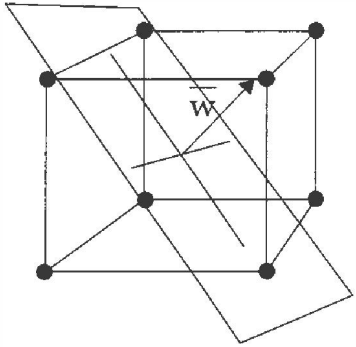

**Figure 2:** The decision surfaces in a binary-valued Naive Bayes network are hyperplanes.

The class of a vertex $\bar{o}$ is $\omega^* = \underset{\omega_k}{\mathrm{argmax}}\left(P\{\omega_k|\overline{O}\}\right)$.

The set of vertices of class $\omega^*$ must form a convex set (because they are in the intersection of (n-1) half spaces, each of the form $r_{jk}(\overline{O}) \geq 0$).

Let $W_{ab}$ be the set of all of the vertices of $\{0, 1\}^n$ such that $P\{\omega_a\} \geq P\{\omega_b\}$ and let $\overline{W}_{ab} = \{0, 1\}^n - W_{ab}$.

The partition $[W_{ab}, \overline{W}_{ab}]$ is called a *dichotomy*. [Saks, 1993] One reasonable question to ask is: How many of the possible dichotomies on $\{0, 1\}^n$ are linearly separable?

**Theorem 2:** The number of dichotomies of the vertices of an n-cube is $2^{2^n}$.

**Theorem 3:** [Muroga, 1971]The number of linearly separable dichotomies of the vertices of an n-cube, $N(n)$, is

$$N(n) \leq \sum_{i=1}^{n} \binom{2^n - 1}{i} < 2\binom{2^n}{n} \cong O\left(2^{n^2}\right)$$

Thus, as the number of dimensions, n, increases, the fraction of dichotomies that a linearly separable becomes vanishingly small.

## 4 NAIVE BAYES CLASSIFICATION WITH M-ARY OBSERVATIONS

There are two strategies for representing extra states. The first is to represent all of the states associated with a single observation along the same dimension. That is, we define an total order on the possible values $\{o_{i1}, ..., o_{ik_i}\}$ for observation $O_i$ and map the values into a set of integers $V_i = \{1, ..., k_i\}$ representing the various states. Instead of defining our decision functions on the vertices of an n-cube, we need to define our function on the product space $V_1 \times V_2 \times ... \times V_n$. In this case, [Duda+Hart, 1973] and others have shown that the decision functions $r_{ab}(\overline{O})$ are polynomials[2]:

$$r_{ab}(\overline{O}) = \sum_{i=1}^{n} \sum_{j=1}^{k_i} w_{ij} O_i^{j-1}$$

The problem with this strategy is that it forces us to impose an ordering on the possible values for each observation. Say that one of our observations is the color of some feature. This color can be one of $\{\text{red, blue, yellow}\}$. Is red < yellow ? blue < red ? In these cases, it makes more sense to use an unordered representation for the states, such as the corners of a simplex. Let $S_n$ be the vertices of the simplex, e.g. $S_n = \{V \in \{0, 1\}^n | |V| = 1\}$. We can map the single observation ("What is the color?") into three binary observations ("Is the color blue?", "Is the color red?", "Is the color yellow?") subject to the constraint that only one of the observations can be true at one time.

The space of allowable observation combinations is a product space $S_{k_1} \times S_{k_2} \times ... \times S_{k_n} \subset \{0, 1\}^\Sigma$ where

$$\Sigma = \sum_{i=1}^{n} k_i.$$

On this space, we can show that the decision functions $r_{ab}(\overline{S})$ are once again linear:

---

2. The derivation of this and similar results is similar to the proof of Theorem 1.

$$r_{ab}(\bar{O}) = w_0 + \sum_{j=1}^{\Sigma} w_i O_i$$

It may seem that this approach is "cheating" by introducing extra dimensions to the problem. If there is no obvious ordering among $k$ variable values, then mapping these values to a set of integers introduces an arbitrary ordering between these variables. The transformation to a simplex, on the other hand, does not introduce any explicit ordering, but does introduce a constraint that all of the variable values be observed.

## 5 OBSERVATION-OBSERVATION DEPENDENCIES

### 5.1 THE EFFECT OF OBSERVATION-OBSERVATION DEPENDENCIES ON R(O)

Observation-observation dependencies make the story more complicated. [Saks, 1993] notes that the decision surfaces for general boolean classification functions are hyperquatrics of the form:

$$r_{ab}(\bar{O}) = \sum_{J \in P(n)} w_J \prod_{i \in J} O_i$$

where $P(n)$ is the power set of $\{1, ..., n\}$.

We can apply a similar trick to equation 2 to model an arbitrary conditional probability distribution $P\{Y_1 | ((Y_2, ..., Y_n), \Omega_j)\}$ when all of the variables $\bar{Y}$ are binary. Once we have shown that we can model this conditional distribution as a polynomial, we will use this transformation to model the conditional distributions in our original classification problem

$$P\{\Omega_j | \bar{O}\} = \left( P\{\Omega_j\} / P\{\bar{O}\} \right) \prod_{O_k \in \bar{O}} P\{O_k | \bar{C}(O_k)\}.$$

Let $\bar{V} = \{0, 1\}^n$. The elements of $\bar{V}$ are of the form $(v_1, v_2, ..., v_n)$. If $\bar{X} = (x_1, x_2, ..., x_n)$ then $X_{2, ..., n} = (x_2, ..., x_n)$

$$P\{Y_1 | Y_{2, ..., n}, \Omega_j\}$$

$$= P\{Y_1 = 0 | (\bar{Y}_{2, ..., n} = \bar{0}), \Omega_j\} \cdot$$

$$\prod_{\bar{v} \in \bar{V}} \left( \frac{P\{Y_1 = v_1 | (\bar{Y}_{2, ..., n} = \bar{v}_{2, ..., n}), \Omega_j\}}{P\{Y_1 = 0 | (\bar{Y}_{2, ..., n} = \bar{0}), \Omega_j\}} \right)^{K(\bar{v}, \bar{Y})}$$

where

$$K(\bar{v}, \bar{Y}) = \prod_{i=1}^{n} (Y_i v_i + (1 - Y_i)(1 - v_i))$$

$K(\bar{v}, \bar{Y})$ is a polynomial in $\bar{Y}$ that is of degree $n$. Let

$$w_{\bar{v}, j} = \log \frac{P\{Y_1 = v_1 | (\bar{Y}_{2, ..., n} = \bar{v}_{2, ..., n}), \Omega_j\}}{P\{Y_1 = 0 | (\bar{Y}_{2, ..., n} = \bar{0}), \Omega_j\}}$$

and $w_{0j} = P\{Y_1 = 0 | (\bar{Y}_{2, ..., n} = \bar{0}), \Omega_j\}$

then

$$\log P\{Y_1 | Y_{2, ..., n}, \Omega_j\} = w_{0, j} + \sum_{\bar{v} \in \bar{V}} w_{\bar{v}, j} K(\bar{v}, \bar{Y}) \quad (4)$$

Now we will use this transformation to transform the conditional distributions our original classification problem. The class probability given an observation is

$$P\{\Omega_j | \bar{O}\} = \frac{P\{\Omega_j\}}{P\{\bar{O}\}} \left( \prod_{O_k \in \bar{O}} P\{O_k | \bar{C}(O_k)\} \right) \quad (5)$$

Where $C(O_k)$ represents the conditional predecessors of $O_k$.

If we substitute $O_k | \bar{C}(O_k)$ for $Y_1 | Y_{2, ..., n}$ in equation (4) and take the log of both sides of (5) we discover:

$$P\{\Omega_j | \bar{O}\} = \log \frac{P\{\Omega_j\}}{P\{\bar{O}\}}$$
$$+ \sum_{O_k \in \bar{O}} \left( w_{0, j, k} + \sum_{\bar{v}_k \in \bar{V}_k} w_{\bar{v}, j, k} K(\bar{v}_k, O_k, \bar{C}(O_k)) \right)$$

Thus, the decision surfaces are polynomials with maximum degree $k = 1 + \max_{O_k \in \bar{O}} |\bar{C}(O_k)|$.

### 5.2 MINIMUM COMPLEXITY OF BELIEF NETWORKS

The combinatorics and computer science literature offer several tantalizing hints at techniques for determining the minimum complexity of a belief network that models a particular dichotomy. For example, a substantial amount of work has been done on bounding both the degree and the density of polynomials for representing a decision function for a particular dichotomy [Saks, 1993]. [Brandman et al, 1990] derive lower bounds on the number of leaves in a decision tree for a given dichotomy, $W$, and the average number of decisions made in any tree for $W$. One of the most surprising results from this literature is the conjecture that as the number of observations go to infinity almost all dichotomies can be represented by polynomials of degree $n/2$.

## 6 DECISION PROBLEMS

Even for simple decision problems, the decision surfaces separating optimal decision regions are more complicated than the decision surfaces of the Naive Bayes model. Consider the influence diagram below.

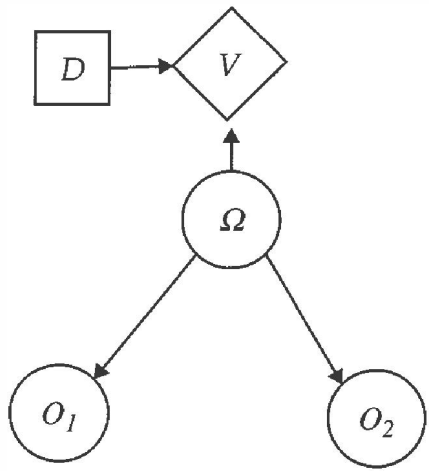

**Figure 3:** A simple influence diagram.

Let $\Omega$ have 4 states, $A$, $B$, $C$, and $D$. $O_1$ and $O_2$ are binary with states $\{0, 1\}$. $D$ has two values: $\{X(OR), \text{not-}X(OR)\}$. $V$ is a binary-valued value function with values $\{0, 1\}$. $P\{A\} = P\{B\} = P\{C\} = P\{D\} = 0.25$.

| | |
|---|---|
| $P\{O_1 = 1 \| \Omega = A\} = 1$ | $P\{O_2 = 1 \| \Omega = A\} = 1$ |
| $P\{O_1 = 1 \| \Omega = B\} = 1$ | $P\{O_2 = 1 \| \Omega = B\} = 0$ |
| $P\{O_1 = 1 \| \Omega = C\} = 0$ | $P\{O_2 = 1 \| \Omega = C\} = 1$ |
| $P\{O_1 = 1 \| \Omega = D\} = 0$ | $P\{O_2 = 1 \| \Omega = D\} = 0$ |

**Figure 4:** Probability distributions for $O_1$ and $O_2$ given $\Omega$

| | |
|---|---|
| $V\{D = X, \Omega = A\} = 0$ | $V\{D = \text{not-}X, \Omega = A\} = 1$ |
| $V\{D = X, \Omega = B\} = 1$ | $V\{D = \text{not-}X, \Omega = B\} = 0$ |
| $V\{D = X, \Omega = C\} = 1$ | $V\{D = \text{not-}X, \Omega = C\} = 0$ |
| $V\{D = X, \Omega = D\} = 0$ | $V\{D = \text{not-}X, \Omega = D\} = 1$ |

**Figure 5:** $V(D, \Omega)$

In this simple problem, the optimal decision policy is $D = X$ whenever $O_1$ xor $O_2$ is true. This is not a linearly separable function in a Bayesian classifier.

## 7 IMPLICATIONS

### 7.1 KNOWLEDGE ACQUISITION

**Conjecture 1:** Unless we have compelling reason to believe that observations are conditionally independent, naive models will not scale. That is, the performance of naive models should decrease with increasing numbers of observations.

One of the implications of Theorem 3 is that the fraction of dichotomies that are linearly separable becomes vanishingly small as the number of dimensions in the problem increases. We conjecture that this effect makes it less likely that the Naive Bayes assumption will be appropriate as the number of dimensions increases, *unless we have compelling reasons to believe that observations are, indeed, conditionally independent.*

**Conjecture 2:** It may be possible to use harmonic analysis of classification dichotomies to identify the minimal required set of observation-observation dependencies in a belief network.

Not too much to say here. It appears that a significant literature exists for bounding the complexity of decision trees and decision functions that compute boolean functions such as class membership. There may be a mapping between this family of techniques and the inherent complexity of a belief network, although this mapping is not obvious.

### 7.2 LEARNING

One testable implication for learning algorithms is Conjecture 1 above.

Another observation, due to Cheeseman [1996], is that the learning community seems to focus on the performance of learning algorithms rather than on the appropriateness of the underlying model. Decision trees strictly dominate hyperplanes in the number of dichotomies that they can represent. However, the strong "hyperplane bias" in a Naive Bayes-based learning algorithm may allow the algorithm to achieve superior performance on small data sets. This hyperplane bias can be though of as a particularly strong prior distribution on the form of the model used to represent a particular joint distribution. Since this strong bias greatly reduces the number of independent parameters in the model, Naive Bayes-based learning algorithms can produce an adequate representation of classification dichotomies using only a few data points. We conjecture that as the amount of data increases, that more general learning algorithms will have superior performance.

## 8 SUMMARY

The primary aim of this paper is to review the well-known linear separability result of Minsky [61] for a new generation of belief network researchers who may not be acquainted with this result. We also illustrate the functional form of decision surfaces for generalizations of the binary Naive Bayes assumption.

## 9 ACKNOWLEDGEMENTS

I would not have written this paper but for the strong encouragement (coercion) of Tom Chavez, Peter Cheeseman, Eric Horvitz, Ross Shachter, David Smith and Alan Wada. This article also benefitted greatly from conversations with Tom Chavez, Peter Cheeseman, Paul Dagum, Denise Draper, Rafe Mazzeo, and Richard Stanley. Thanks also to the anonymous reviewers for catching some major typos.